\pdfoutput=1

\documentclass[hyphens,11pt]{article}

\usepackage[final]{acl}

\usepackage{times}
\usepackage{latexsym}

\usepackage[T1]{fontenc}

\usepackage[utf8]{inputenc}

\usepackage{microtype}
\usepackage{graphicx}
\usepackage{subcaption}
\usepackage{color, colortbl}
\usepackage{tablefootnote}
\usepackage{enumitem}  
\usepackage{lipsum}

\usepackage{lipsum}

\newcommand\nusacrowd{\includegraphics[width=13pt]{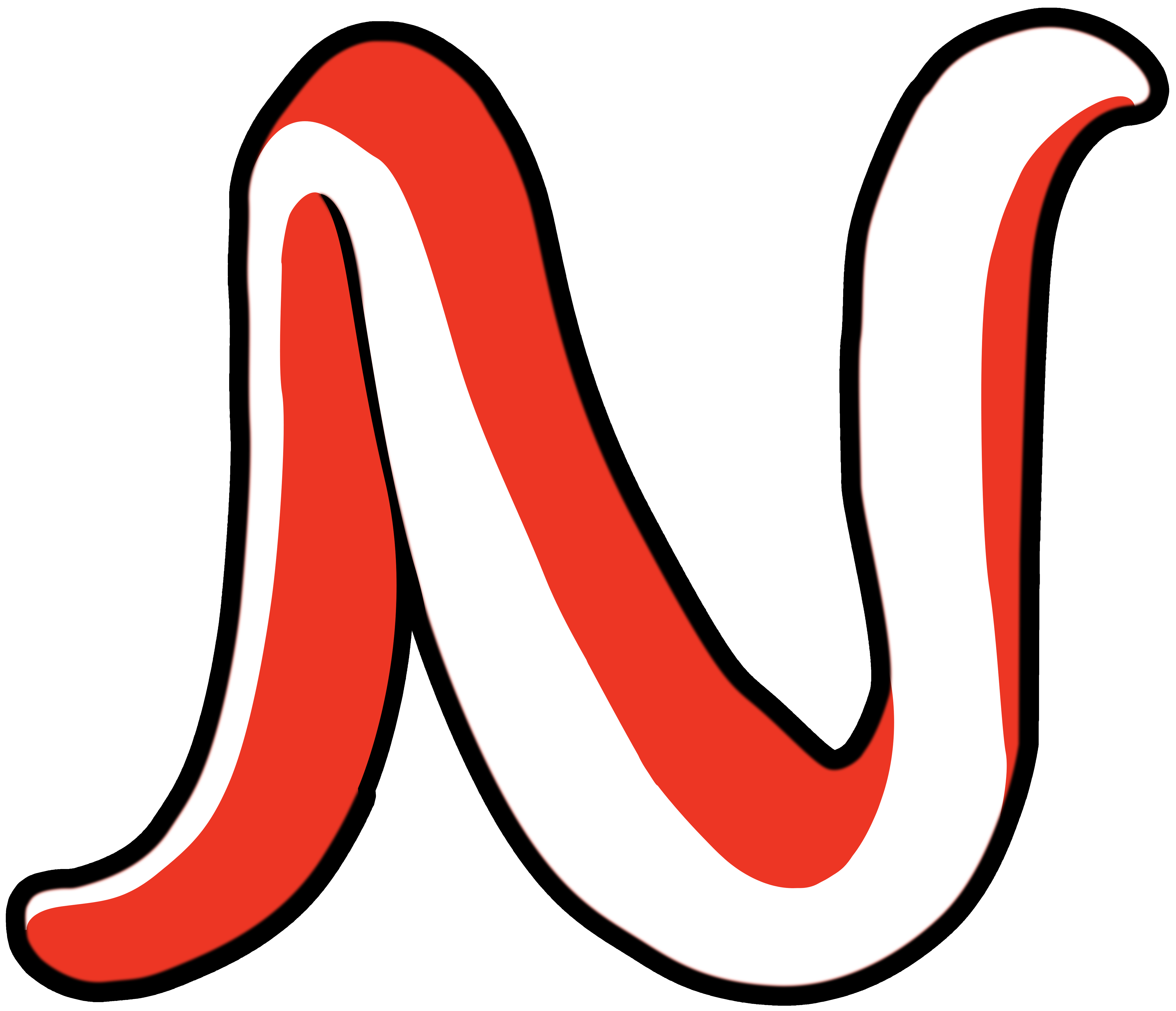}}
\newcommand\indflag{\includegraphics[width=13pt]{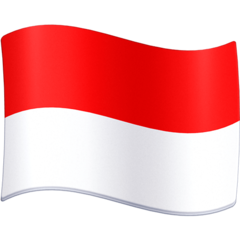}}

\newcommand{\equalsign}{\footnotemark[1]\hspace{0.1cm}}

\title{NusaCrowd: A Call for Open and Reproducible NLP Research \\ in Indonesian Languages}

\author{
  Samuel Cahyawijaya\thanks{\hspace{0.2cm}Equal contribution.}\hspace{0.12cm}, Alham Fikri Aji\equalsign, Holy Lovenia\equalsign, \\
  \textbf{Genta Indra Winata}\equalsign, \textbf{Bryan Wilie}\equalsign, \textbf{Rahmad Mahendra},  \textbf{Fajri Koto}, \\
  \textbf{David Moeljadi}, \textbf{Karissa Vincentio}, \textbf{Ade Romadhony}, \textbf{Ayu Purwarianti} \\
  IndoNLP 
  \\
}

\begin{document}
\maketitle
\begin{abstract}



At the center of the underlying issues that halt Indonesian natural language processing (NLP) research advancement, we find data scarcity. Resources in Indonesian languages, especially the local ones, are extremely scarce and underrepresented. Many Indonesian researchers refrain from publishing and/or releasing their dataset. Furthermore, the few public datasets that we have are scattered across different platforms, thus makes performing reproducible and data-centric research in Indonesian NLP even more arduous. 
Rising to this challenge, we initiate the first Indonesian NLP crowdsourcing effort, NusaCrowd.
NusaCrowd strives to provide the largest datasheet aggregation with standardized data loading for NLP tasks in all Indonesian languages.
By enabling open and centralized access to Indonesian NLP resources, we hope NusaCrowd can tackle the data scarcity problem hindering NLP progress in Indonesia and bring NLP practitioners to move towards collaboration.
\end{abstract}

\section{What is \nusacrowd~NusaCrowd?}
\label{sec:what-is-nusacrowd}


Natural language processing (NLP) resources in Indonesian languages, especially the local language ones, are extremely scarce and underrepresented in the research community.
This introduces bottlenecks to Indonesian NLP research, restraining it from opportunities and hindering its progress.
In response to this issue, several Indonesian and NLP communities have sourced various types of datasets to also be available in Indonesian languages~\citep{winata2022nusax,koto-etal-2021-evaluating,cahyawijaya2021indonlg,koto-etal-2021-indobertweet,mahendra-etal-2021-indonli,aji-etal-2021-paracotta,wilie2020indonlu,abdul2020mega,ilmania2018absa,pan2017cross}. However, a significant mass of the local resources is scattered across different platforms~\citep{etsa2018pengaruh, apriani2016pengaruh,dewi2020combination}, and a lack of access to public datasets still persists~\cite{aji-etal-2022-one}.

\begin{figure}
    \centering
    \includegraphics[width=\linewidth, trim={24cm 8cm 24cm 8cm}, clip]{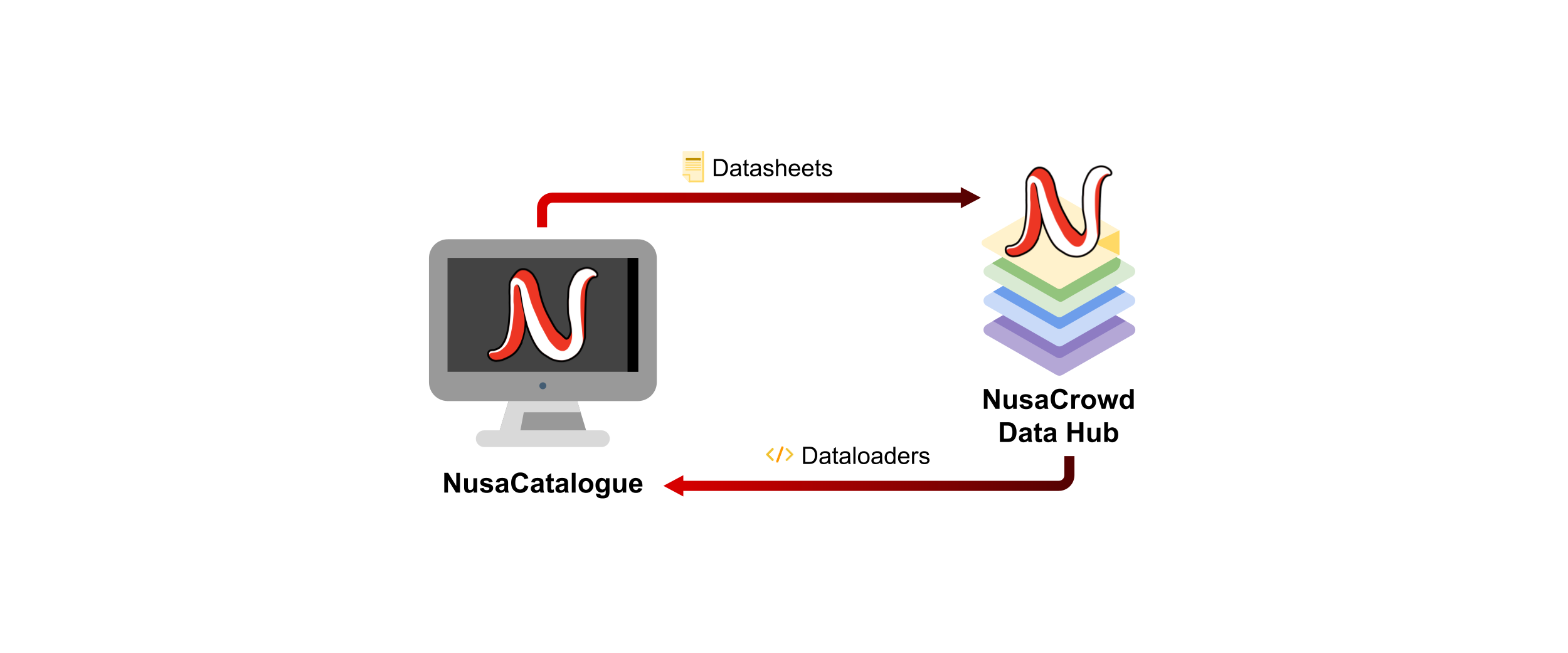}
    \caption{
    Open access to the datasheets collected is provided through \textbf{NusaCatalogue}, and the dataloader scripts to retrieve the resources are implemented in \textbf{NusaCrowd Data Hub}.}
    \label{fig:nusacatalogue-nusacrowd}
\end{figure}

To address this vital problem, inspired by other open collaboration projects~\cite{orife2020masakhane,nekoto2020participatory,dhole2021nlaugmenter,zaid2022masader,mcmillan2022documenting,srivastava2022beyond,fries2022bigbio,gehrmann2022gemv2},
we take a step and initiate NusaCrowd, a joint movement to collect and centralize NLP datasets in Indonesian and various Indonesia's local languages, and engage the linguistics community in collaboration. 

\begin{figure*}[t]
    \centering
    \includegraphics[width=\linewidth, trim={15cm 5cm 15cm 5cm}, clip]{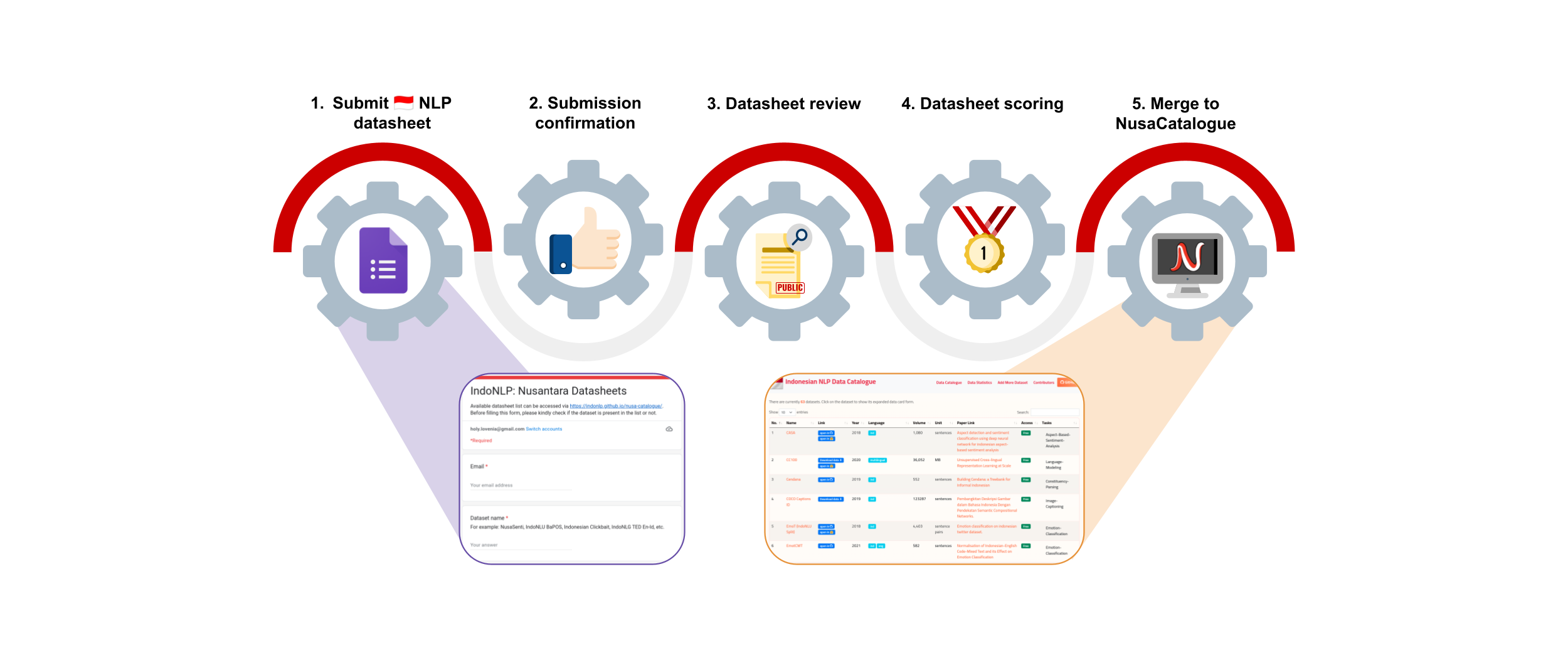}
    \caption{The outline of public Indonesian NLP datasheet submission to NusaCatalogue.}
    \label{fig:flow-public-datasheet}
\end{figure*}

Powered by the collective effort of our contributors, NusaCrowd aims to increase the accessibility of these datasets and promote reproducible research on Indonesian languages through three fundamental facets: 1) Curated public corpora datasheet sourcing\footnote{\url{https://indonlp.github.io/nusa-catalogue/}}, 2) Open-access centralized data hub\footnote{\url{https://github.com/IndoNLP/nusa-crowd}}, and 3) Promoting private-to-public data access. We maintain the quality of the contributions, both the consolidation efforts and the programmatic means, by enforcing a quality control with a mix of automatic and manual evaluation schemes.
NusaCrowd is currently open for contribution, the movement is held from 25 June 2022 to 18 November 2022.
Let's bring Indonesian NLP research one step forward together.


\section{Contributing in \nusacrowd~NusaCrowd}
\label{sec:contributing-in-nusacrowd}

Together, contributors in NusaCrowd drive Indonesian NLP forward by developing a multi-faceted solution to improve data accessibility and research reproducibility. To assist our widespread open collaboration and collective progression, we formulate three main ways of contributing in NusaCrowd in the following sections, each corresponds to a fundamental aspect in NusaCrowd's main objective.



\subsection{Submit public \indflag~NLP datasheet}
\label{sec:nusa-catalogue}

We encourage contributors to register the datasheets of public datasets on \textbf{NusaCatalogue}\footnote{\url{https://indonlp.github.io/nusa-catalogue/}} by submitting them through an online form at~\url{https://forms.gle/31dMGZik25DPFYFd6}. NusaCatalogue is a public datasheet catalogue website, inspired by~\citet{zaid2022masader}, in which we list all datasets collected in NusaCrowd. We build NusaCatalogue to improve the discoverability of Indonesian NLP datasets and to assist users in searching and locating Indonesian NLP datasets based on their metadata.

A datasheet is a dataset metadata which contains various information about the dataset, including but not limited to: dataset name, original resource URL, relevant publication, supported tasks, and dataset licence. The datasheet will be reviewed and scored within a week or two. The contribution point calculation for the public Indonesian NLP corpora datasheet is based on three criteria: 1) whether the relevant dataset is previously public or not, 2) dataset quality, and 3) language rarity. Details on the scoring mechanism will be explained in \S\ref{sec:public-corpora}.

Once the datasheet passes the review, we will notify the responsible contributor
and list the approved dataset's datasheet on NusaCatalogue and also reported on NusaCrowd data hub task list\footnote{\url{https://github.com/orgs/IndoNLP/projects/2}} so its dataloader could be implemented. The complete flow of how to submit the public Indonesian NLP corpora datasheet is shown in Figure~\ref{fig:flow-public-datasheet}.

\begin{figure*}[t!]
    \centering
    \includegraphics[width=\linewidth, trim={15cm 5cm 15cm 5cm}, clip]{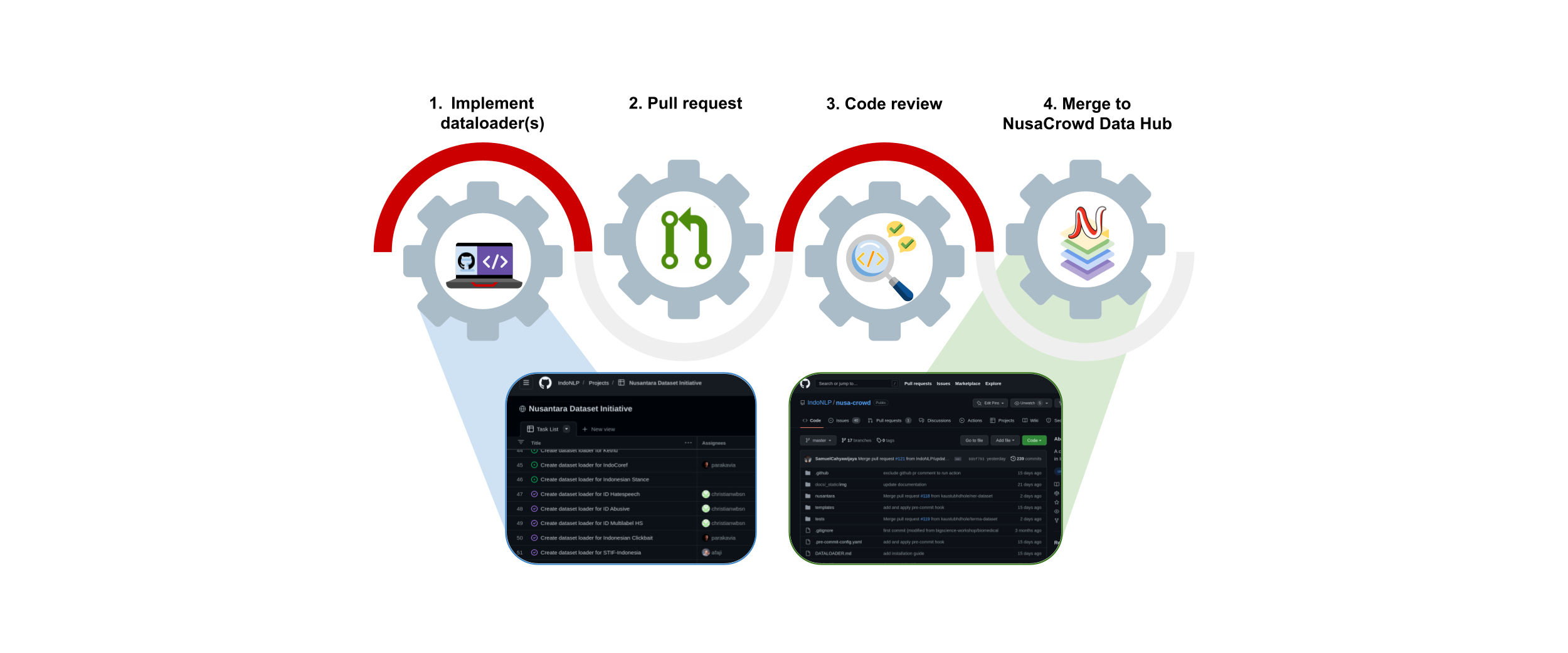}
    \caption{The outline of dataloader implementation for NusaCrowd data hub.}
    \label{fig:flow-dataloader-implementation}
\end{figure*}

\subsection{Implement dataloader(s) for \nusacrowd~NusaCrowd data hub}

A large-scale centralized data hub has to be equipped with the capability of a simple and standardized programmatic data access that spans across diverse resources, regardless of their separate hosting locations, distinct data structures or formats, and different configurations. For this purpose, building NusaCrowd data hub requires a few key elements: datasheet documentation (\S\ref{sec:nusa-catalogue}), task schema standardization to support common NLP tasks, and dataloader implementation. While the datasheets become the backbone of NusaCrowd and standardized task schemas compose the skeleton, the dataloader implementation is the heart of NusaCrowd data hub.

Each dataset requires a specific dataloader script tailored to its source task type, structure, and configuration to enable easy loading and enforce interoperability. To centralize all of the Indonesian NLP resources, NusaCrowd needs a large number of proper dataloaders to be implemented. Therefore, we invite all collaborators to contribute through creating these dataloaders via NusaCrowd's GitHub\footnote{\url{https://github.com/IndoNLP/nusa-crowd}}.

Firstly, a contributor can view the task list in NusaCrowd Github project, then choose the dataset they want to implement by assigning themself to the related issue. Afterwards, the contributor can start setting up the environment needed for development. To help with the dataloader implementation, we provide a template script specifying all the parts the contributor needs to complete, and task schemas for common NLP tasks, e.g., knowledge base, question answering, text classification, text-to-text, text pairs, question answering, and more. We also equip NusaCrowd's repository with several working dataloader scripts that the contributor can check for examples. To fill in the details of the dataset in the dataloader, such as its source URL or its publication, the contributor can refer to the corresponding datasheet recorded in NusaCatalogue.

The contributor can ensure that their dataloader is implemented correctly through a manual inspection, a direct attempt of execution, and a unit test provided in the repository. We also encourage the contributor to tidy up their code accordingly with our formatter before they make a pull request to submit their changes. The implemented dataloader then will go through a code review process by NusaCrowd maintainers. If an adjustment in the code is required, the maintainer will request some changes and provide their feedback on a comment on the respective pull request, so the contributor will be able to improve the dataloader accordingly. Once two maintainers give their approvals, the dataloader will be merged to NusaCrowd repository. The flow overview of the dataloader implementation is depicted in Figure \ref{fig:flow-dataloader-implementation}. A comprehensive guide for contributing through dataloader implementation can be accessed here\footnote{\url{https://github.com/IndoNLP/nusa-crowd/blob/master/DATALOADER.md}}.

\subsection{Provide information on private \indflag~NLP datasets}

\begin{figure*}
    \centering
    \includegraphics[width=\linewidth, trim={15cm 4cm 15cm 5cm}, clip]{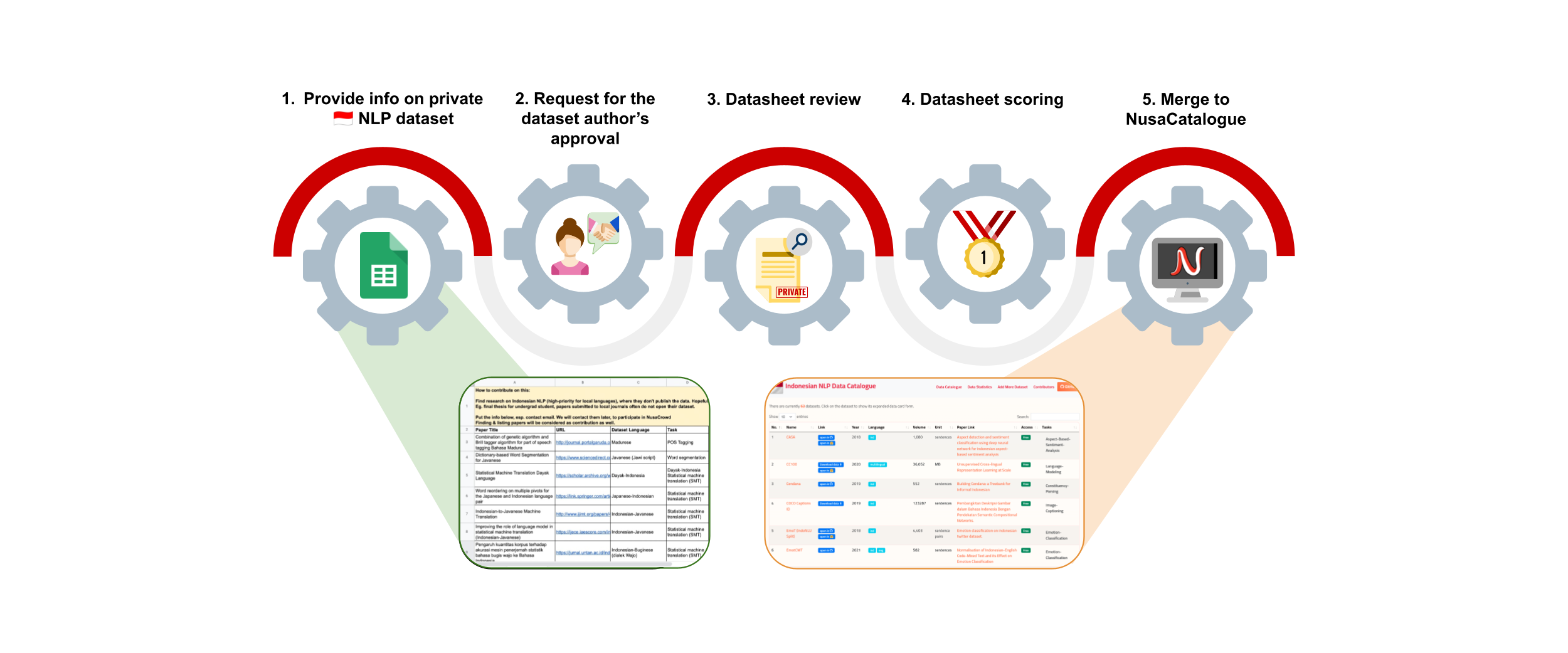}
    \caption{The outline of private Indonesian NLP dataset listing in NusaCatalogue.}
    \label{fig:flow-private-dataset}
\end{figure*}

Many studies in Indonesian NLP use private datasets, which in turn hampers the research reproducibility. Therefore, the last method to contribute is to list research papers of Indonesian NLP in which the data is not shared publicly. We then contact the author to participate in open data access and ask for their approval to include the dataset in NusaCrowd. Contribution points for the authors that release their data to public will follow the scoring defined in \S\ref{sec:non-public-corpora}.
The paper lister will also be awarded with a contribution point. 

As far as we know, there is no data or analysis yet on why local researchers prefer not to share their dataset publicly. Some reasons that we are aware of include: 1) not accustomed to open data and open research, 2) restricted by the university or funding policy, or 3) keep the data private as property. By listing research with private datasets, we can additionally ask the authors about their consideration to improve our understanding on this matter. Steps to provide information on private Indonesian NLP datasets are illustrated in Figure \ref{fig:flow-private-dataset}.




\section{Contribution Point}
\label{sec:contrib}


To support fairness and transparency for all of our contributors, we establish a scoring system of which co-authorship eligibility will be decided from. To be eligible as a co-author in the upcoming NusaCrowd publication, a contributor needs to earn at least 10 contribution points. The score is aggregated from all the contributions made by the contributors. In order to earn contribution points, we introduce three different methods to contribute (for the method details, see \S\ref{sec:contributing-in-nusacrowd}): 1) submitting public Indonesian NLP corpora datasheet,
2) implementing dataloader(s) for NusaCrowd data hub, and 3)
provide information on non-public Indonesian NLP datasets. The point for each type of contribution is described in the following paragraphs.

\begin{figure*}[t]
    \centering
    \includegraphics[width=1.0\linewidth, trim={7.5cm 3cm 9cm 3cm}, clip]{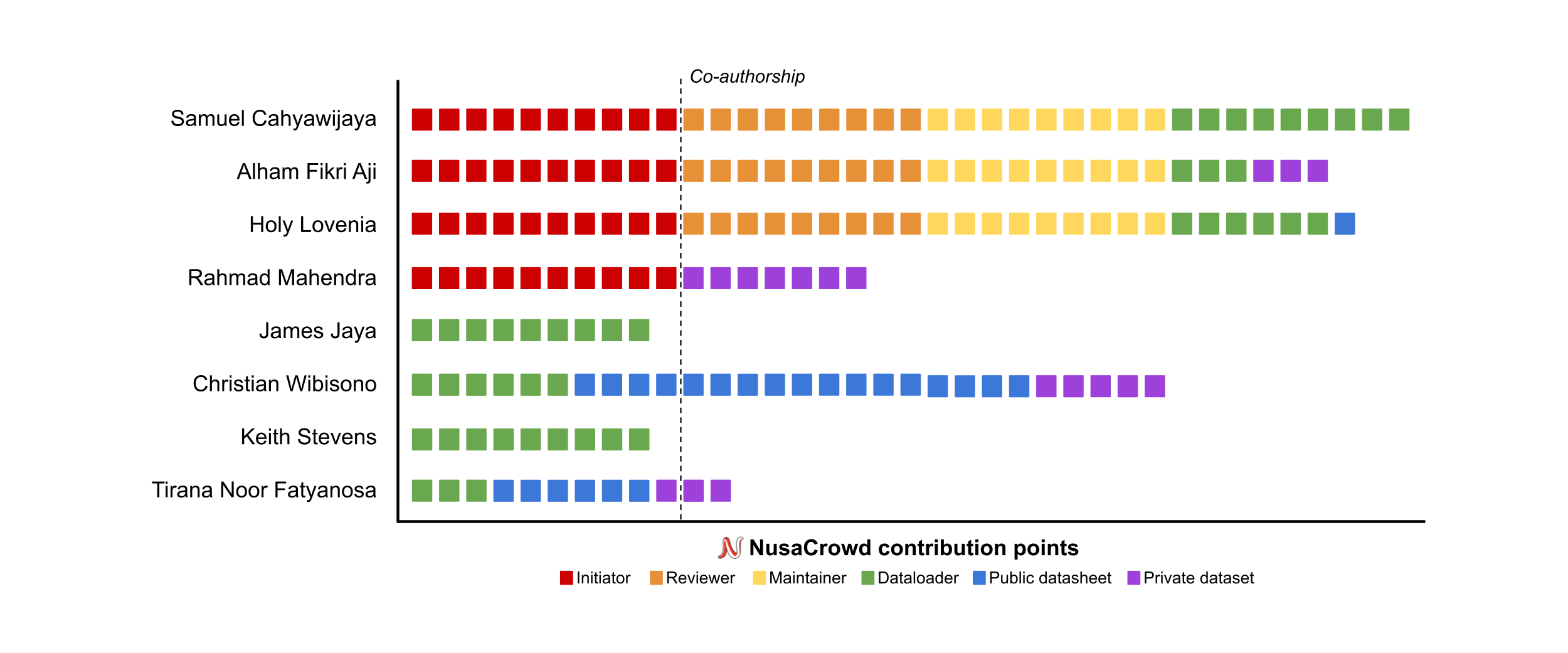}
    \caption{A glance at the contribution matrix used for recapitulation. The contributor list is clipped for simplicity.}
    \label{fig:contribution-matrix}
\end{figure*}

\subsection{Public \indflag~NLP corpora datasheet}
\label{sec:public-corpora}

A contributor can help to register public NLP corpora in NusaCrowd. For any datasheet listed, the contributor is eligible for +2 contribution point as a referrer. To support the development of local language datasets, we provide additional contribution points according to the rarity of the dataset language. Specifically, a contributor of any Sundanese (sun), Javanese (jav), or Minangkabau (min) dataset, will receive +2 contribution points, while a contributor of any other local language dataset will be granted +3 contribution points.

In addition, to encourage more diverse NLP corpora, we provide additional +2 contribution points for tasks that are considered rare. Based on our observation, we find that the common NLP tasks in Indonesian languages include: machine translation (MT), language modeling (LM), sentiment analysis (SA), and named entity recognition (NER). All other NLP tasks are considered rare and are eligible for the +2 contribution points. Lastly, we also notice that publicly available Indonesian NLP corpora involving another modality (e.g., speech or image) are very scarce, for instance: speech-to-text or automatic speech recognition (ASR), text-to-speech (TTS) or speech synthesis, image-to-text (e.g., image captioning), text-to-image (e.g., controllable image generation), etc. To encourage more coverage over these data, we will give additional +2 contribution points for the relevant datasheets submitted.



We understand that dataset quality can vary a lot. To support fairness in scoring datasets with different qualities, for any dataset that does not pass a certain minimum standard, 50\% penalty will be applied. This penalty affects any dataset that is collected with: 1) crawling without manual validation process, 2) machine or heuristic-rule labelled dataset without manual validation, and 3) machine-translated dataset without manual validation.


\subsection{Implementing \nusacrowd~dataloader }
\label{sec:dataloder}

A contributor can help to implement a dataloader for any dataset listed on the task list in NusaCrowd GitHub project (see \S\ref{sec:dataloder}). As a rule of thumb, one dataloader implementation is generally worth 3 contributions points. However, there are some exceptions where a dataloader can be worth more. The contribution points will be counted once the respective pull request is merged to \texttt{master}.

\subsection{Finding and opening private \indflag~NLP datasets}
\label{sec:non-public-corpora}

Contributors can help to list research papers introducing non-public Indonesian NLP dataset. For every private dataset listed, the corresponding contributor will be eligible for +1 contribution point. While for the original author of the dataset, if the author agrees to make the dataset publicly available, the author will be eligible for +3 contribution points. Note that we might request the author to clean up their data until  it is proper for public release (i.e. formatting, consistency, or additional filtering). In addition to the contribution points obtained from publicly releasing the dataset, the author is also eligible for additional points from the datasheet listing, as mentioned in \S\ref{sec:public-corpora}, when the datasheet is submitted to NusaCatalogue.

\subsection{Other ways to contribute in \nusacrowd}

Other than the previously mentioned contributions methods, we also open for other forms of contribution, subject to NusaCrowd's open discussion. To get more details on the open discussion, please join our Slack and Whatsapp group (see \S\ref{sec:call-for-participation}).

\subsection{Contribution point recapitulation}

The total contribution point for all contributors will be recapped every week by a maintainer, and a contribution matrix will be published and updated on a weekly basis. The contribution matrix and more detailed information to the contribution point can be accessed on the following link\footnote{\url{https://docs.google.com/spreadsheets/d/e/2PACX-1vS3Kbi9s3o_V-lyFRHeONOI7jFnMlUswqKj-D6cpgiSYOSxbijC4DIrjAstqxj-H-EI6I2lFhhyKe5s/pubhtml}}. The final score will be recapped in November and the finalized contribution matrix (see Figure \ref{fig:contribution-matrix} for an example) will be published along with the research paper.

\begin{figure*}[t!]
    \centering
    \includegraphics[width=1.0\linewidth, trim={27.25cm 13.5cm 25cm 13.5cm}, clip]{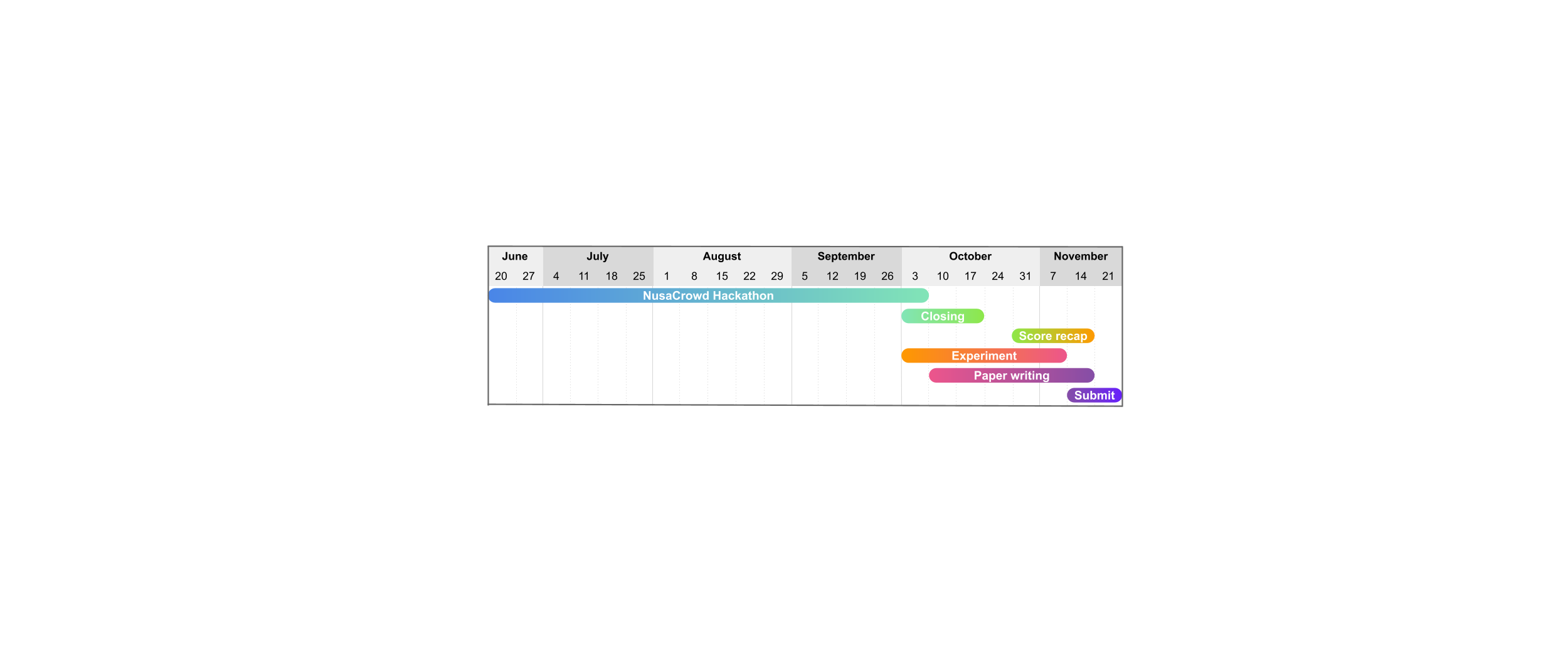}
    \caption{Timeline of the NusaCrowd movement. The datasheet collection and dataloader implementation start from 25 June 2022 to 2 October 2022. Then NusaCrowd will continue with experiments, paper writing, and point recapitulation. The paper will be submitted to a top-level computational linguistics conference, ACL 2023.}
    \label{fig:gantt-chart}
\end{figure*}

\begin{figure*}[t!]
	\begin{subfigure}{0.33\linewidth}
        \centering{\includegraphics[width=0.95\columnwidth]{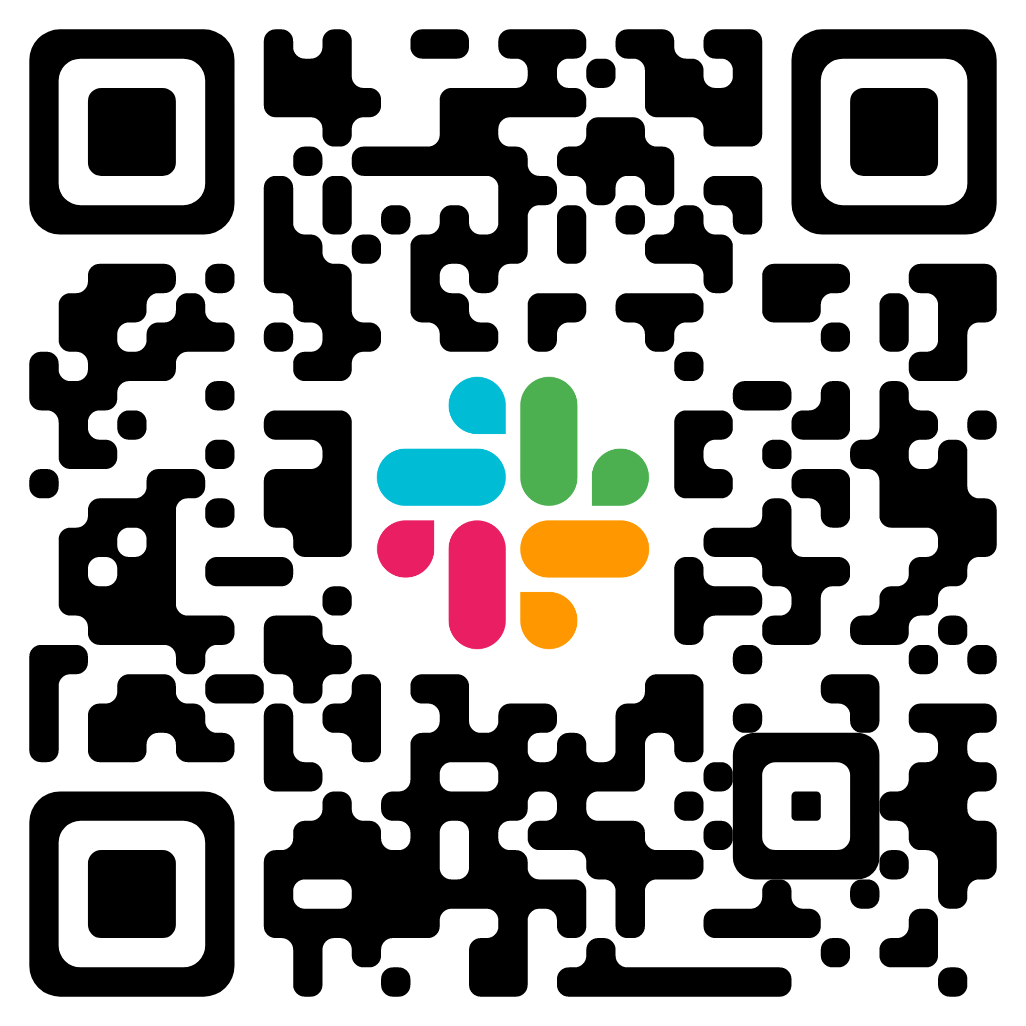}}
	\end{subfigure}
	\begin{subfigure}{0.33\linewidth}
        \centering{\includegraphics[width=0.95\columnwidth]{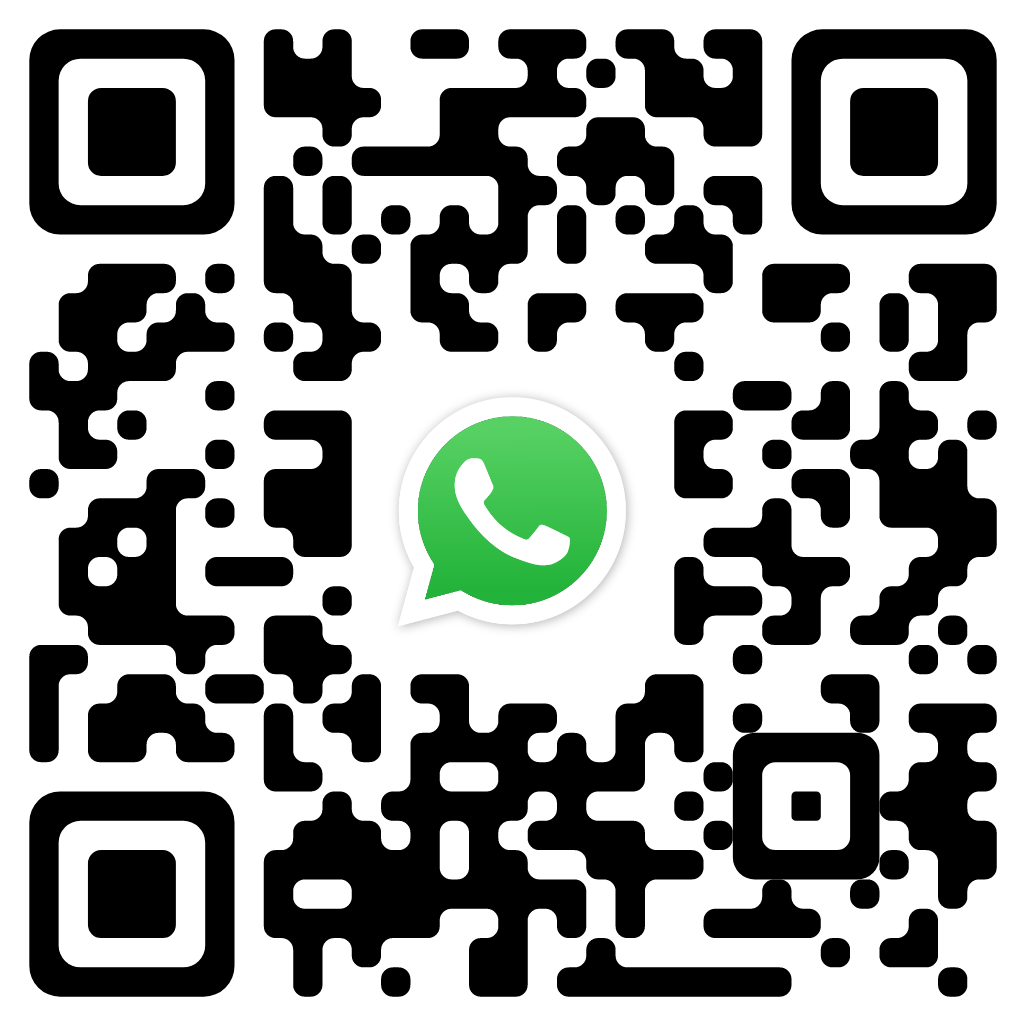}}
	\end{subfigure}
	\begin{subfigure}{0.33\linewidth}
        \centering{\includegraphics[width=0.95\columnwidth]{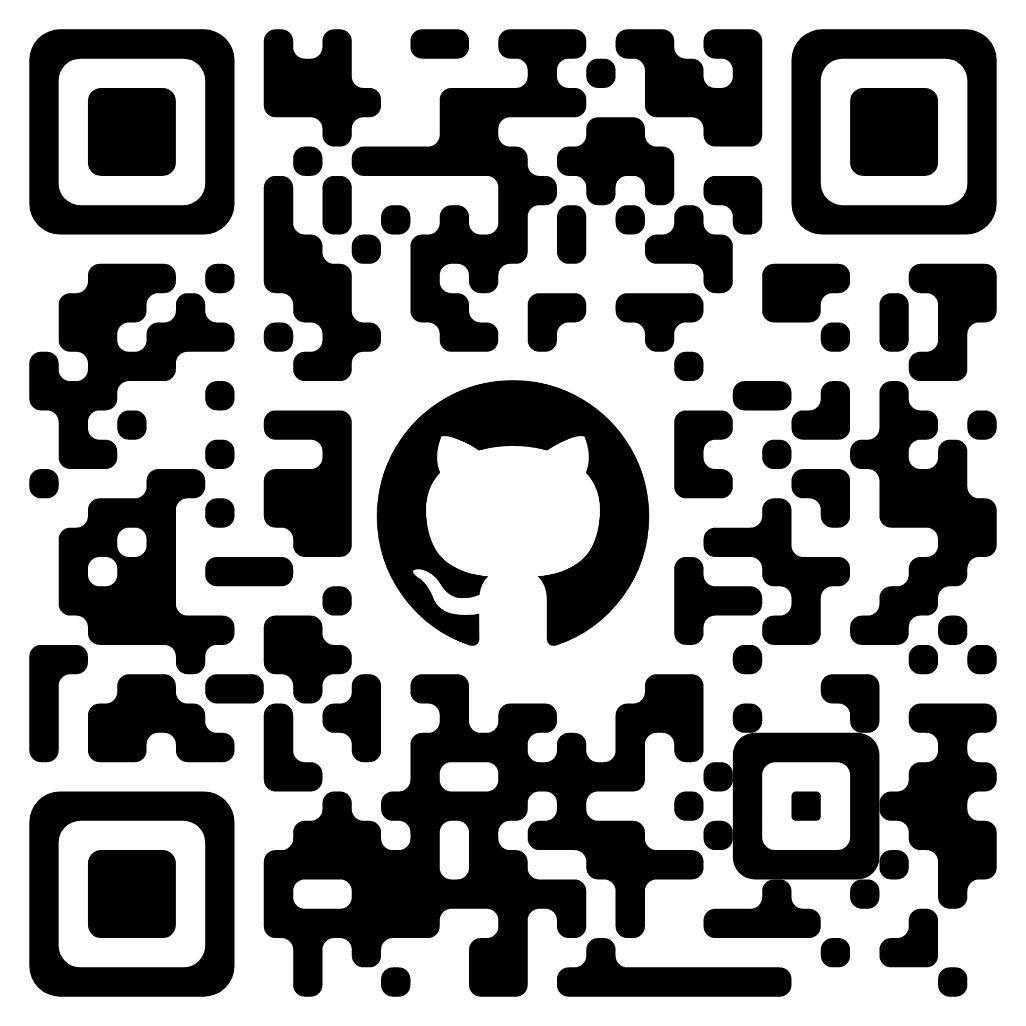}}
    \end{subfigure}
    \caption{Join NusaCrowd's Slack (\url{https://join.slack.com/t/nusacrowd/shared_invite/zt-1b61t06zn-rf2rWw8WFZCpjVp3iLXf0g}), Whatsapp group (\url{https://chat.whatsapp.com/Jn4nM6l3kSn3p4kJVESTwv}), and Github (\url{https://github.com/IndoNLP/nusa-crowd}).}
    \label{fig:join_us}
\end{figure*}

\section{Dataset Licensing and Ownership}

NusaCrowd does not make a clone or copy the submitted dataset. The owner and copyright holder will remain to the original data owner. All data access policy will follow the original data licence without any modification from NusaCrowd. NusaCrowd simply downloads and reads the file from the original publicly available data source location when creating the dataloader, and, in addition, the NusaCatalogue datasheet also directly points to the original data site and publication.

\section{Timeline}

The current NusaCrowd movement is started from 25 June 2022 and will be closed on 18 November 2022. The registration of datasheets and dataloaders will be completed on 2 October 2022. From October onwards, we will focus more on preparing extension and set of experiments to show the benefit of having NusaCrowd platform. In addition, the contribution point for each contributors (see Figure \ref{fig:contribution-matrix} for an example) and the research paper will also be finalized by early November, followed by the final submission of the paper to the Association of Computational Linguistics (ACL) 2023 conference. The detailed phase for the paper development is shown in Figure~\ref{fig:gantt-chart}.










\section{Summary}

NLP resources in Indonesian languages, especially the local ones, are extremely low-resource and underrepresented in the research community. There are multitudes factors causing this limitation. Here we solve this problem by initiating the largest Indonesian NLP crowd sourcing efforts, NusaCrowd. In the spirit of fairness, openness, and transparency; NusaCrowd comes with various ways to contribute towards openness and standardization in Indonesian NLP, while at the same time, introducing a scoring mechanism that provides equal chance for all contributors to show the best out of their contributions. We hope that, NusaCrowd can bring a new perspective to all Indonesian NLP practitioners to focus more on openness and collaboration through code and data sharing, complete documentation, and community efforts.

\section{Call for participation}
\label{sec:call-for-participation}

We invite all Indonesian NLP enthusiasts to participate in NusaCrowd. For any inquiry and further information, contributors can join our community channel on Slack or Whatsapp Group (see Figure~\ref{fig:join_us}). Let's work together to advance the progress of Indonesian language NLP!~\indflag ~\indflag~\indflag


\section*{Acknowledgements}
Thank you for all initiators, without whom NusaCrowd initiative would never be possible, and salute to all contributors for the amazing efforts in NusaCrowd.

\bibliography{custom,anthology}
\bibliographystyle{acl_natbib}




\end{document}